\documentclass{INTERSPEECH2023}
\pdfoutput=1
\usepackage{amssymb}
\usepackage{tabularx}
\usepackage{tikz}
\usetikzlibrary{positioning, arrows.meta, calc, shapes.misc, matrix}

\interspeechcameraready

\title{Allophant: Cross-lingual Phoneme Recognition with Articulatory Attributes}

\name{Kevin Glocker$^1$\thanks{This work was partially supported by the German Federal Ministry of Education and Research (BMBF) under funding nr. 16DHBKI032}, Aaricia Herygers$^1$, Munir Georges$^{1,2}$}

\address{
  $^1$AImotion Bavaria, Technische Hochschule Ingolstadt, Germany\\
  $^2$Intel Labs, Germany}
\email{firstname.lastname@thi.de}

\begin{document}

\maketitle

\begin{abstract}

This paper proposes Allophant, a multilingual phoneme recognizer. It requires only a phoneme inventory for cross-lingual transfer to a target language, allowing for low-resource recognition.
The architecture combines a compositional phone embedding approach with individually supervised phonetic attribute classifiers in a multi-task architecture. We also introduce Allophoible, an extension of the PHOIBLE database. When combined with a distance based mapping approach for grapheme-to-phoneme outputs, it allows us to train on PHOIBLE inventories directly.

By training and evaluating on 34 languages, we found that the addition of multi-task learning improves the model's capability of being applied to unseen phonemes and phoneme inventories.
On supervised languages we achieve phoneme error rate improvements of 11 percentage points (pp.) compared to a baseline without multi-task learning.
Evaluation of zero-shot transfer on 84 languages yielded a decrease in PER of 2.63 pp.\ over the baseline.

\end{abstract}


\section{Introduction}
Speech technologies such as automatic speech recognition (ASR) have greatly improved in recent years.
However, despite the increase in research interest \cite{scharenborg2017building, magueresse2020_lowresource-review, kamper2020multilingual, yu2020acoustic,yadav-sitaram-2022-survey, van2022feature}, for many low-resource languages these technologies are still not available or perform poorly. Such languages may be endangered or lack a ``stable orthography" \cite{BESACIER2014under-resourced}.
Similarly, many models experience difficulties in recognizing regional or non-native accented speech \cite{huang2012cross, feng2021quantifying, herygers2023bias}, for which there is often little data available.

To reduce the amount of training data needed for these languages, various techniques (e.g., probabilistic transcriptions \cite{hasegawa2017under-resourced}) and benchmarks \cite{nguyen2020zero-resource} have been proposed.
To avoid the dependence on training data in the target language, architectures have been introduced that allow zero-shot phoneme recognition on unseen languages using only their phoneme inventories \cite{li_universal_2020, li21f_interspeech}.
Both architectures use allophone-to-phoneme mappings (i.e., variations of phonemes that retain word meanings but are generally articulated in distinct contexts \cite{ladefoged2014course}).

In the ASR2K architecture \cite{Li_Metze_Mortensen_Black_Watanabe_2022b}, multilingual acoustic and pronunciation models are trained jointly without any supervision.
They successfully use pre-trained self-supervised learning (SSL) models, such as the multilingually pre-trained XLS-R \cite{Babu_Wang_Tjandra_2022, Li_Metze_Mortensen_Black_Watanabe_2022b}.
Additionally, multi-task learning with articulatory attributes has been shown to improve character error rates in Mandarin \cite{Lee2019MultitaskLF} and cross-lingual phoneme error rates \cite{glocker-georges-2022-hierarchical}.

Following the examples set by Allosaurus \cite{li_universal_2020}, a ``multilingual allophone system", and AlloVera \cite{mortensen-etal-2020-allovera}, a ``multilingual allophone database", we thus present our multilingual phoneme recognition architecture called ``Allophant" and phoneme database called ``Allophoible". Allophant combines an allophone layer and phonetic composition \cite{Li_Li_Metze_Black_2021, Li_Metze_Mortensen_Black_Watanabe_2022b} with multi-task learning for cross-lingual phoneme recognition.

We describe the Allophant architecture in Section~\ref{sec:allophant}.
In Section~\ref{sec:allophoible} we present Allophoible, an extension of the PHOIBLE database \cite{phoible} that includes articulatory attributes for additional phoneme segments such as allophones.
We describe our experiments in Section~\ref{sec:evaluation} and present and discuss the results in Section~\ref{sec:results} before concluding in Section~\ref{sec:conclusion}.

\section{Allophant Architecture}\label{sec:allophant}

\vspace{-1em}
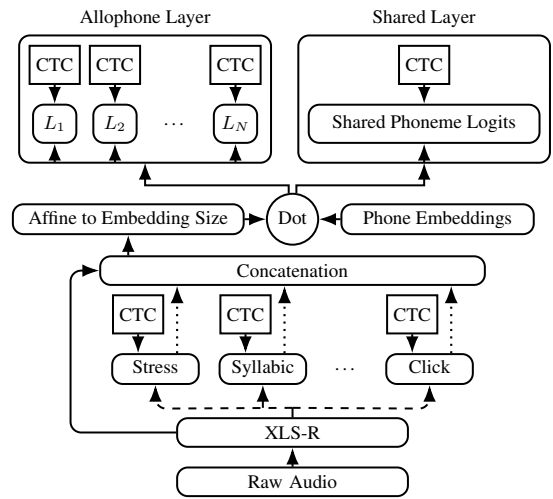
\begin{figure}[h]
\begin{center}
\begin{tikzpicture}[scale=0.8, transform shape]
    \begin{scope}[
        local bounding box=asrschematic,
        box/.style={draw, text centered, rounded corners, minimum height=1.5em},
        ctc/.style={draw, text centered, minimum height=2em, minimum width=2.25em},
        line width=0.75pt
    ]
        \begin{scope}[
            local bounding box=attributes,
            attribute/.style={box, minimum width=4.5em, minimum height=1.5em, font=\vphantom{Ag}},
        ]
            \node[attribute](stress){Stress};
            \node[attribute, right=1em of stress.east](syllabic){Syllabic};
            \node[minimum width=2em, right=1em of syllabic.east](others){\dots};
            \node[attribute, right=1em of others.east](click){Click};
            \foreach \classifier in {stress, syllabic, click} {
                \node[ctc, anchor=south west](\classifier_ctc) at ($(\classifier.north) + (-2.25em, 1em)$){CTC};
                \draw[-Latex](\classifier_ctc) -- (\classifier_ctc |- \classifier.north);
            }
        \end{scope}
        \node[box, minimum width=12em, below=1.75em of attributes](wav2vec){XLS-R};
        \node[box, minimum width=12em, text width=10em, below=1em of wav2vec](asrinputs){Raw Audio};

        \node[box, minimum width=20em, above=0.5em of attributes](concat){Concatenation};
        \coordinate(above_concat) at ($(concat.north) + (0, 1.9em)$);
        \node[box, minimum width=12em, minimum height=1.5em, left=2.5em of above_concat, anchor=east](phone_out){Affine to Embedding Size};
        \node[circle, draw, minimum width=2.5em, font=\vphantom{Ag}](dot) at (above_concat){Dot};
        \node[box, minimum width=10em, minimum height=1.5em, anchor=west, right=2.5em of above_concat](compose){Phone Embeddings};
        connections
        \draw[-Latex] (asrinputs) -- (wav2vec);
        \coordinate(attribute_center_below) at ($(attributes.south) - (0, 1.25em)$);
        \draw (wav2vec) -- (attribute_center_below);
        \foreach \attribute in {stress, click} {
            \draw[-Latex, dashed, rounded corners](attribute_center_below) -- ($(\attribute.south) - (0, 1.25em)$) -- (\attribute.south);
        }
        \draw[-Latex](syllabic |- attribute_center_below) -- (syllabic);

        \coordinate(concat_center_below) at ($(concat.south) - (0, 1em)$);
        \foreach \attribute in {stress, syllabic, click} {
            \coordinate(\attribute_top_right) at ($(\attribute.north) + (1.125em, 0)$);
            \draw[-Latex, dotted, thick] (\attribute_top_right) -- (\attribute_top_right |- concat.south);
        }

        \draw[-Latex] (phone_out) -- (dot);
        \draw[-Latex] (compose) -- (dot);
        \draw[-Latex] (concat.north -| phone_out) -- (phone_out);
        \coordinate(outside_left) at ($(concat.west) - (1.5em, 0)$);
        \draw[-Latex, rounded corners] (wav2vec.west) -- (outside_left |- wav2vec.west) -- (outside_left) -- (concat);

        \begin{scope}
            \coordinate(above_phones) at ($(above_concat) + (0, 5em)$);
            \begin{scope}[local bounding box=allophones, language/.style={box, minimum width=2.25em, minimum height=2em, anchor=south}]
                \node[language, left=1.8em of above_phones, anchor=east](l_n){$L_N$};
                \node[minimum width=2.25em, left=0.8em of l_n.west](other_langs){\dots};
                \node[language, left=0.8em of other_langs](l_2){$L_2$};
                \node[language, left=0.8em of l_2.west](l_1){$L_1$};
                \foreach \i/\language in {1/l_1, 2/l_2, 3/l_n} {
                    \node[ctc, above=1em of \language](allo_ctc_\i){CTC};
                    \draw[-Latex](allo_ctc_\i) -- (\language);
                    \draw[-Latex]($(\language.south) - (0, 1em)$) -- (\language);
                }
                \coordinate(allo_bottom_left) at (allo_ctc_1.west |- l_1.south west);
                \coordinate(allo_bottom_right) at (allo_ctc_3.east |- l_n.south east);
                \draw[rounded corners] ($(allo_ctc_1.west) - (0.5em, 0)$) -- ($(allo_ctc_1.north west) + (-0.5em, 0.5em)$) -- ($(allo_ctc_3.north east) + (0.5em, 0.5em)$) -- ($(allo_bottom_right) + (0.5em, -1em)$) -- ($(allo_bottom_left) - (0.5em, 1em)$) -- ($(allo_ctc_1.west) - (0.5em, 0)$);
            \end{scope}
            \node[above=0.05em of allophones, align=center]{Allophone Layer};

            \begin{scope}[local bounding box=shared_module, anchor=south]
                \node[box, minimum width=12em, minimum height=2em, right=0.8em of above_phones](shared){Shared Phoneme Logits};
                \node[ctc, above=1em of shared](shared_ctc){CTC};
                \draw[-Latex] (shared_ctc) -- (shared);
                \draw[-Latex]($(shared.south) - (0, 1em)$) -- (shared);
                \coordinate(shared_top_left) at (shared.east |- shared_ctc.north east);
                \coordinate(shared_top_right) at (shared.west |- shared_ctc.north west);

                \draw[rounded corners] ($(shared.south west) - (0.5em, 0)$) -- ($(shared_top_right) + (-0.5em, 0.5em)$) -- ($(shared_top_left) + (0.5em, 0.5em)$) -- ($(shared.south east) + (0.5em, -1em)$) -- ($(shared.south west) - (0.5em, 1em)$) -- ($(shared.south west) - (0.5em, 0)$);
            \end{scope}
            \draw[-Latex] ($(dot.north) + (0.2em, 0)$) -- ($(dot.north) + (0.2em, 0.4em)$) -| (shared_module.south);
            \node[above=0.05em of shared_module, align=center]{Shared Layer};
        \end{scope}
        \draw[-Latex] ($(dot.north) - (0.2em, 0)$) -- ($(dot.north) + (-0.2em, 0.4em)$) -| (allophones.south);
    \end{scope}
\end{tikzpicture}
\vspace{-1.4em}
\end{center}
\caption{Illustration of the Allophant architecture, including an allophone layer \protect{\cite{li_universal_2020}}, phone composition \protect{\cite{li21f_interspeech}} and hierarchical multi-task connections \protect{\cite{glocker-georges-2022-hierarchical}}. Multi-task models use the dashed connections and hierarchical multi-task models also use the dotted connections. ``Shared'' models use the shared layer while all other models use the allophone layer instead.}\label{fig:architecture}
\end{figure}

The proposed Allophant architecture and its evaluated variants are illustrated in Figure~\ref{fig:architecture}.
As in previous work on zero-shot phoneme recognition \cite{Li_Metze_Mortensen_Black_Watanabe_2022b, Xu_Baevski_Auli_2021}, we finetune a model trained with an SSL objective.
We choose the 300 million parameter version of the cross-lingual speech representation model (XLS-R), which was pre-trained on 128 languages \cite{Xu_Baevski_Auli_2021}.
XLS-R was found to outperform other SSL models for this task \cite{Li_Metze_Mortensen_Black_Watanabe_2022b}.

As the phoneme classifier, we use the compositional architecture from \cite{li21f_interspeech} and apply it to the PHOIBLE feature set \cite{phoible} as motivated in Section~\ref{sec:allophoible}.
In this architecture, phone embeddings are computed by summing the embeddings of their attribute values.
We designate an embedding to represent each possible value of the 35 articulatory attributes in addition to a special ``blank'' attribute, which represents blanks for connectionist temporal classification (CTC) training \cite{Graves2006ConnectionistTC, li21f_interspeech}.

Each attribute can have a value of ``+'' if it exists, ``-'' if it does not, or ``0'' if it is impossible in combination with other attributes of a phoneme. The only exception for this is the stress attribute, which is only ``-'' or ``0'' in PHOIBLE.
Since we do not consider tonal languages (e.g., Mandarin) in this paper, the tone attribute was unused and therefore removed from training and inference.
Complex phoneme segments such as some diphthongs, affricates, and multiple articulations contain attribute contours such as ``+,-,-'' instead of only a single value.
For simplicity, we only use the first attribute from complex contours for the composition.

Out of 107 attribute values, six are not used by any phonemes in the training data and are replaced by zero vectors to not affect inference.
The missing attribute values are [+retractedTongueRoot], [+advancedTongueRoot], [+epilaryngealSource], [+raisedLarynxEjective], [+loweredLarynxEjective], and [+click].
We then compute phone logits with the scaled dot product \cite{Vaswani2017AttentionIA, li21f_interspeech} between the output hidden representation from the final XLS-R layer and the composed phone or blank embeddings.
At training time, phoneme probabilities are computed by additionally passing the phone logits through an allophone layer \cite{li_universal_2020, li21f_interspeech, Li_Metze_Mortensen_Black_Watanabe_2022b}.
The resulting language-specific phoneme sequences are then trained with CTC loss.

Following our previous work \cite{glocker-georges-2022-hierarchical}, we additionally explicitly supervise articulatory attribute classifiers for all 35 attributes that are also used for embedding composition.
In this architecture, CTC loss is computed for each attribute sequence individually and minimized together with the phoneme CTC loss during training.
In contrast to embedding composition, we use the full complex contours in label sequences for attribute classifiers.
As a result, attribute sequences may differ in length.

To analyze the interactions between embedding composition, multi-task learning and the allophone layer, we test five variations of the architecture.
All variants use phoneme composition.
Two different architectures that are trained only on phoneme recognition without multi-task learning are used as baselines.
Our first baseline is closest to the compositional model first introduced in \cite{li21f_interspeech} and used as the acoustic model in \cite{Li_Metze_Mortensen_Black_Watanabe_2022b}.
We also consider a simplified form of the proposed architecture without an allophone layer to better understand the interaction of multi-task learning with phoneme composition on its own.
In variants using this ``Shared'' layer instead we compose embeddings for the union of all phonemes that occur in the training languages instead of all allophones during training and optimize it directly.
This configuration is comparable to shared phoneme models used in previous work \cite{li_universal_2020}, where a feed-forward projection layer was used instead of embedding composition to compute phoneme logits \cite{Xu_Baevski_Auli_2021, glocker-georges-2022-hierarchical}.

In addition, we evaluate two more multi-task architectures to identify how the attribute classifiers impact phoneme recognition.
``Multi-Task Shared'' is the multi-task analogue to ``Baseline Shared'' for direct comparison without the allophone layer.
``Multi-Task Hierarchy'' adds the hierarchical connection between attribute classifiers and the phoneme classifier as in \cite{glocker-georges-2022-hierarchical}.
In this architecture, probability distributions from attribute classifiers are concatenated with the acoustic model outputs from XLS-R before passing the resulting vector through the affine transformation to embedding size.

\section{Allophoible: Allophone Inventory}\label{sec:allophoible}
Allophoible\footnote{https://github.com/Aariciah/allophoible/releases/tag/v1.0.0} (version 1.0) is an open-access inventory that builds upon PHOIBLE 2.0 \cite{phoible}, which contains phoneme information for $\approx$2200 languages.
We provided phoneme attributes for diacritics and phonemes that were not included as phonemes in PHOIBLE 2.0 but were included as allophones or in eSpeak NG \cite{espeak} (version 1.51).
This allows us to use the compositional architecture from \cite{li21f_interspeech} with PHOIBLE allophone mappings and compose phoneme embeddings for eSpeak NG phonemes.

\begin{table}[t]
\caption{Zero-shot transfer results on UCLA including average PERs, AERs and the variances of PERs between languages.}\label{tab:ucla}
\centering
\resizebox{\columnwidth}{!}{%
\begin{tabular}{l*{3}{c}}
\toprule
Name & PER & PER $\sigma^2$ & AER\\
\midrule
Baseline (15k steps) & 57.01\% & 374.35 & --\\
Baseline Shared & 48.25\% & 250.59 & --\\
Multi-Task Shared & 46.05\% & 250.34 & 19.52\%\\
Multi-Task & \textbf{45.62\%} & 229.06 & 19.44\%\\
Multi-Task Hierarchy & 46.09\% & 230.28 & \textbf{19.18\%}\\
\bottomrule
\end{tabular}%
}
\end{table}

\begin{figure}[h]
    \centering
    \includegraphics[width=1.0\columnwidth]{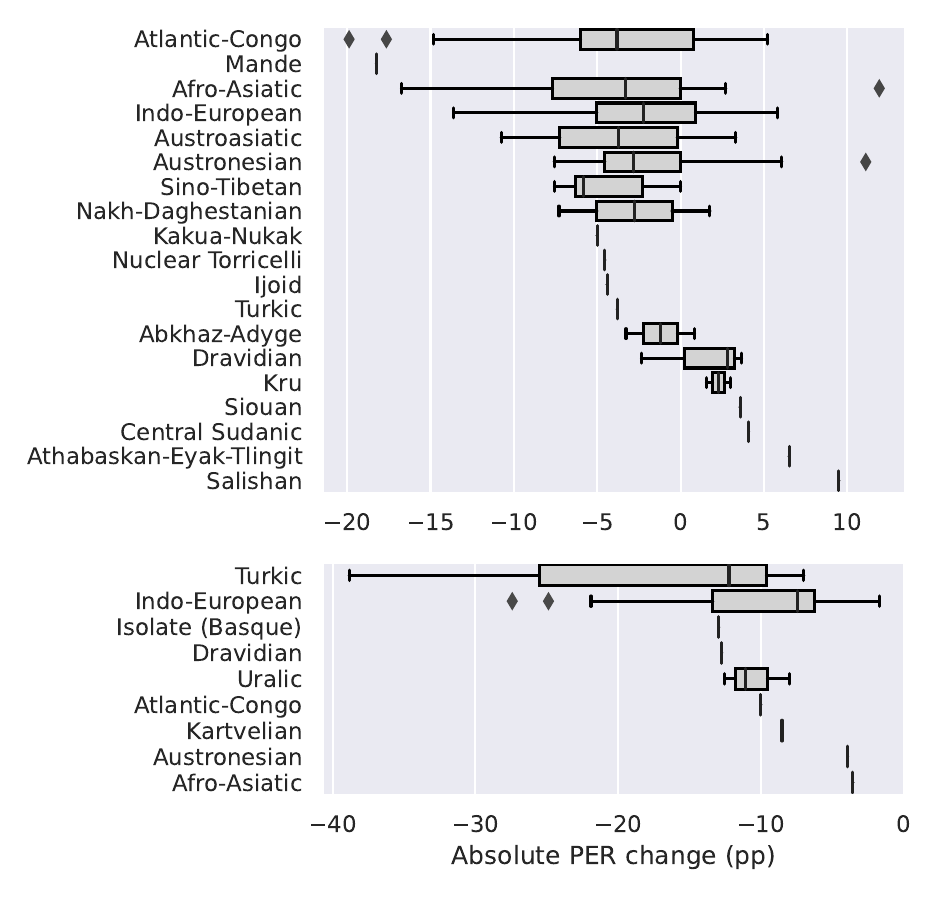}
    \caption{Absolute changes in PER on UCLA (top) and CV (bottom) by language family (Glottolog \protect{\cite{glottolog}}) from ``Baseline Shared'' to ``Multi-Task'' learning in percentage points.}
    \label{fig:relative_improvements}
\end{figure}

\section{Experiments}\label{sec:evaluation}

\subsection{Datasets}
We selected a subset of at most 800,000 utterances per language from 34 eSpeak-supported, non-tonal languages from the Mozilla Common Voice (CV) corpus \cite{ardila-etal-2020-common} (version 10.0).
In total, our training set consisted of 4628h of speech.
For validation and testing we used the CV development and test sets and filtered some utterances as described in Section~\ref{sec:processing}.

Cross-lingual transfer was evaluated on the phonetically transcribed utterances from the first release of the UCLA Phonetic Corpus \cite{li2021multilingual}.
It contains 5,509 utterances from 95 low-resource languages, of which 11 occurred in our training set.
Since evaluation on these languages would not be zero-shot, they were removed. Their ISO 639-3 codes are: ``ben'', ``ces'', ``dan'', ``ell'', ``eus'', ``fin'', ``hin'', ``hun'', ``lit'', ``mlt'', and ``nld''.

Furthermore, from the remaining 84 languages, 17 were included in the pre-training data for XLS-R.
While the model being aware of the acoustic properties of these languages could benefit phoneme recognition after fine-tuning, we did not find any significant effect (Welch's T-test: $t(22.7) = 0.64, p = 0.528$ for ``Multi-Task").
Their ISO 639-3 language codes are: ``abk", ``afr", ``asm", ``guj", ``hau", ``haw", ``heb", ``hrv", ``hye", ``isl", ``kan", ``khm", ``lav", ``lug", ``mal", ``mya", ``yue". For testing, we use the phoneme inventories specified in the corpus to compose phoneme embeddings for each language.

\subsection{Data and Processing}\label{sec:processing}

We used the allophone inventories available in PHOIBLE \cite{phoible} to provide allophone mappings for the architectures using an allophone layer.
This is applied to all languages in the training set.
Since the phoneme inventories in the database differ to varying degrees from the grapheme-to-phoneme output, we mapped phonemes to their closest equivalents in each inventory.

All phonemes that are already in the target inventory were retained.
Similarly to previous work \cite{Xu_Baevski_Auli_2021}, we then mapped the remaining phonemes to the target phonemes with the lowest attribute Hamming distance.
For simplicity, we only considered the first attribute in contours of complex segments for this mapping.
Finally, we prevented e.g., diphthongs being mapped to single vowels by splitting complex segments whenever the number of segments between pairs of matched phonemes differ.
After splitting, each sub-segment was mapped individually.
Using this approach, we covered approximately 79\% of PHOIBLE inventories for each language compared to 17.5\% without it.

To provide a phoneme transcription of the CV data we used eSpeak NG, which was shown to outperform other grapheme-to-phoneme models \cite{Xu_Baevski_Auli_2021}.
Characters in eSpeak NG that were not International Phonetic Alphabet characters were manually replaced.
For eSpeak NG, we constructed inventories from the union of all phonemes that occur in the transcriptions of each language.
In all three CV subsets (i.e., train, validation, and test) we removed the segments with mixed scripts (e.g., Latin character text in Tamil sentences) as eSpeak NG treats them as being in a different language.
This would lead to a mix in phoneme inventories (e.g., English phonemes in the Tamil inventory).
To ensure that evaluation results are not skewed based on our mapping, we test on the unmapped reference transcriptions. For this, we computed embeddings for all phonemes that occur in the training and test set for all languages at test time.

\subsection{Training}
Allophant was implemented in Python with PyTorch \cite{NEURIPS2019_9015} and Torchaudio \cite{yang2021torchaudio}. Our implementation is open-source\footnote{https://github.com/kgnlp/allophant}.
We use the pre-trained XLS-R model provided by Hugging Face with the transformers library \cite{wolf-etal-2020-transformers, Babu_Wang_Tjandra_2022}.
We use a dropout rate of 0.2 after the final XLS-R output layer.
The attribute embedding size for composition is set to 640 as in previous work \cite{li21f_interspeech}.
We use dynamic batching where batches are constructed from as many raw audio frame sequences as can fit in an at most 16,000,000 element matrix, including padding.

Following \cite{Babu_Wang_Tjandra_2022}, we use language upsampling with $\alpha = 0.5$ for more diverse multilingual training batches.
We use transformer-style warmup \cite{Vaswani2017AttentionIA} but we keep the learning rate constant for 10,000 steps after linearly increasing it for 2,500 before starting the decay.
This change and step numbers are inspired by the schedule from \cite{Xu_Baevski_Auli_2021}.
Also following their work, we keep the feature encoder of XLS-R frozen. However, we finetune all other layers for the entire training process.
We finetuned our models for 30,000 updates on single NVIDIA A100 instances with 40gb of memory.
The average training duration was $\approx$60 hours.
Since not all utterances are seen after 30,000 updates, we ensured that batches were sampled with the same random seed.

\section{Results \& Discussion}\label{sec:results}
\begin{figure*}
    \centering
    \includegraphics[width=1.0\textwidth]{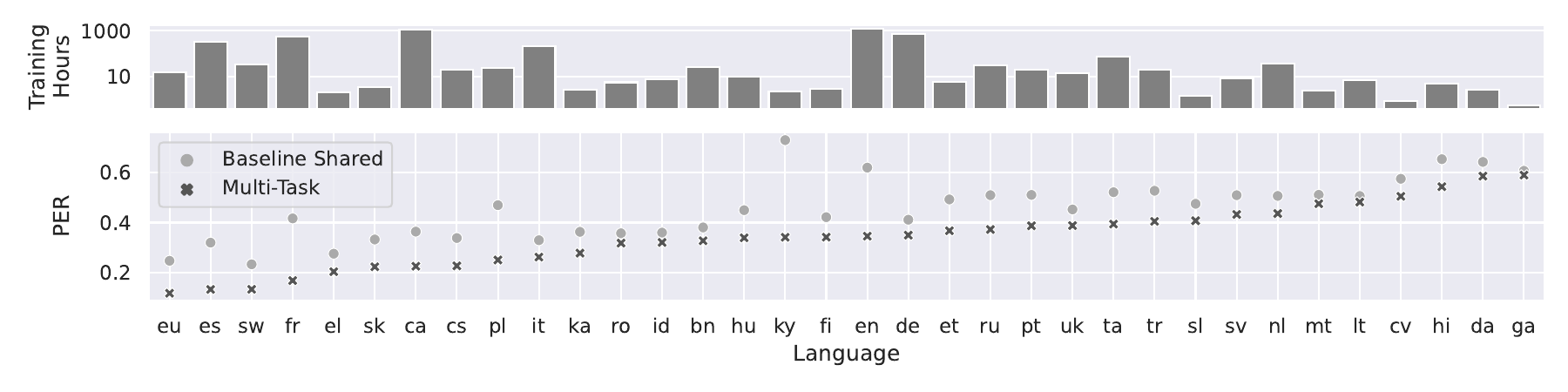}
    \caption{``Baseline Shared" and ``Multi-Task"  PERs for CV test languages with corresponding hours of training data on a log scale.}
    \label{fig:common_voice_languages}
\end{figure*}

Results for zero-shot transfer are presented in Section~\ref{sec:ucla-results} and supervised languages analyzed in Section~\ref{sec:common-voice-results}.
We use average attribute error rate (AER) for evaluation.
It is computed by calculating the average of individual attribute error rates as in \cite{glocker-georges-2022-hierarchical}.
The ``Baseline'' model was evaluated after 15,000 (15k) steps instead of 30,000  to allow for a fairer comparison since it started to overfit to the training data. PERs after 30,000 steps increased by 7.5 pp.\ and 3.3 pp.\ on UCLA and CV respectively.

\subsection{Zero-Shot Transfer on the UCLA Phonetic Corpus}\label{sec:ucla-results}

\begin{table}[t]
\caption{CV test set results for languages in the training data.}\label{tab:common-voice}
\centering
\resizebox{\columnwidth}{!}{%
\begin{tabular}{l*{3}{c}}
\toprule
Name & PER & PER $\sigma^2$ & AER\\
\midrule
Baseline (15k steps) & 46.95\% & 133.31 & --\\
Baseline Shared & 45.35\% & 140.17 & --\\
Multi-Task Shared & 41.20\% & 153.31 & 8.88\%\\
Multi-Task & \textbf{34.34\%} & 149.25 & \textbf{8.36\%}\\
Multi-Task Hierarchy & 34.35\% & 128.86 & 8.56\%\\
\bottomrule
\end{tabular}%
}
\end{table}

Evaluation results of the zero-shot capabilities of our model on the UCLA Phonetic Corpus are listed in Table~\ref{tab:ucla}.
``Baseline Shared" performs substantially better than ``Baseline" by 8.76 percentage points (pp.) PER.
This suggests that the combination of composition and allophone layer does not generalize as well.

An explanation for this potential overfitting is that the allophone layer allows the model to maximize logits of allophones of a phoneme even if its attributes do not represent the acoustic signal well enough.
The small difference between ``Baseline'' and ``Baseline Shared'' on the training data presented in Section~\ref{sec:common-voice-results} supports this hypothesis.
Potentially, the initialization of the attribute embeddings lead to some allophones that have a high feature Hamming distance to the corresponding phoneme to start with high logits.
Since only the most probable allophone is selected in the allophone layer via max pooling \cite{li_universal_2020}, overfitting to initially high scoring allophones is possible.

Both non-hierarchical ``Multi-Task'' models outperform our strongest baseline model (``Baseline Shared'').
The ``Multi-Task Shared'' model lowers the PER by 2.2 pp.\ and ``Multi-Task'' by 2.63 pp., showing the overall benefit of explicitly supervising attribute classifiers to improve the generalization of unseen languages.
Compared to ``Baseline'', adding multi-task learning decreases the PER substantially by 11.39 pp.\ without the overfitting issues.
We suspect that attribute level supervision signals indirectly help stabilize the attribute embeddings in the composition layer.
This might prevent them from diverging too much from underlying attributes to maximize specific allophones.

We did not encounter the same benefits of using hierarchical multi-task learning as in our previous work \cite{glocker-georges-2022-hierarchical}.
With a PER difference of 0.47 pp.\ it performs almost identically.
This might be since in contrast to a projection layer with independent weights for each phoneme, composed phone embeddings do not suffer from the same sparsity and improvements of attribute embeddings for one phoneme also benefits others.
This sharing of attribute-level information could only be achieved by the hierarchical connection in \cite{glocker-georges-2022-hierarchical}.
On the attribute level, we found that AERs are within $\approx0.3$ of each other across architecture variations.
While the hierarchical models reach the lowest AERs, the difference is smaller than in our previous work \cite{glocker-georges-2022-hierarchical} likely due to the same effects we observe on a phoneme level.

In contrast to \cite{glocker-georges-2022-hierarchical}, where a feed-forward layer was used instead of embedding composition and no correlation between PER and AER was found ($r^2 = 0.016$), we find a moderate correlation for ``Multi-Task'' ($r^2 = 0.679$). This shows, how acoustic representations that benefit attribute classification are more likely to also directly improve phoneme recognition with the compositional approach.

\subsection{Common Voice}\label{sec:common-voice-results}

Results on the CV tests sets for languages in the training data are shown in Table~\ref{tab:common-voice}.
We find overall very similar patterns in how models perform relative to each other as on UCLA.
When comparing the baselines, the difference between the ``Baseline'' and ``Baseline Shared'' models of 1.6 PER is much smaller than on UCLA.
This provides further evidence for its tendency to overfit to the training data and languages.

As in the zero-shot results, both multi-task models outperform the baselines. Compared to the baselines, the impact of multi-task learning is larger on CV than on unseen languages. ``Multi-Task Shared'' outperforms ``Baseline Shared'' by 4.15 pp.\ PER. Our ``Multi-Task'' model lowers PER substantially by 11 pp.\ over ``Baseline Shared'' and is our best model overall. This shows that multi-task learning benefits the embedding composition layers through improved acoustic representations both cross-lingually and on supervised languages.

We find that ``Multi-Task Hierarchical'' yields almost identical PERs as ``Multi-Task'', with only a decrease in PER variance across languages. This shows that the hierarchical connection does not benefit supervised languages either, likely for the same reasons outlined in Section~\ref{sec:ucla-results}.
For higher resource languages, additional resources such as phoneme n-gram language models \cite{Xu_Baevski_Auli_2021} have been shown to decrease PERs further. Investigating this is left for future work.
Correlation between PER and AER is lower on CV than UCLA for ``Multi-Task'' ($r^2 = 0.571$).

A more detailed comparison of error rates across languages between ``Baseline Shared'' and ``Multi-Task'' model alongside hours of training data can be seen in Figure~\ref{fig:common_voice_languages}. ``Multi-Task'' reduces PERs to varying degrees for every language in CV.
We improve recognition the most on Kyrgyz (ky, 38.8 pp.), English (en, 27.4 pp.), French (fr, 24.9 pp.), and Polish (pl, 21.9 pp.).
To further analyze the difference in recognition performance, the relative changes are shown by language family in Figure \ref{fig:relative_improvements}. Recognition of the three Turkic languages improved the most on average, mainly through Kyrgyz and Turkish. It is followed by the 22 Indo-European languages, Basque, and Tamil. The latter is the single Dravidian language in our training set.

In UCLA, Atlantic-Congo is the language family with the most improvements, likely due to the PER improvements of 10 pp.\ in Swahili (sw). Afro-Asiatic languages likely also benefit from the 3.6 pp.\ PER improvement of Maltese (mt).
Furthermore, PER improvements for Austroasiatic and Austronesian languages can be explained by the improved modeling of the single languages from these families in our CV training set.
In some language families, PER also increases after Multi-Task learning, which requires further investigation in future work. In particular, for the single Salishan language in UCLA, PER rises by almost 10 pp. over ``Baseline Shared''.

\section{Conclusion}\label{sec:conclusion}
We proposed Allophant, a multilingual phoneme recognizer that can be applied zero-shot to low-resource languages.
Its architecture combines phonetic embedding composition with multi-task articulatory attribute classifiers, improving phoneme recognition across both unseen and supervised languages.
We also introduced Allophoible, an extension of the PHOIBLE database with attributes for all of its allophones.
Together with a mapping scheme of phonemes from eSpeak NG to PHOIBLE inventories, this allowed us to train models with a previously introduced attribute composition and allophone layer on 34 languages.

Additionally, our multi-task learning model outperformed our strongest baseline without it by 2.63 pp.\ PER on cross-lingual transfer to 84 languages of the UCLA phonetic corpus.
Furthermore, our multi-task model outperformed the baseline by
11 pp.\ PER on the supervised languages from the CV corpus.
We showed that multi-task learning compliments the attribute embedding composition architectures by improving their generalization to unseen inventories and reducing overfitting.

A limitation of this work is that phoneme inventories with attributes must be available for the target languages. Future work could expand on inferring inventories for unseen languages \cite{Li_Metze_Mortensen_Black_Watanabe_2022a}.
More work is needed to investigate the effects of the Allophant architecture on the recognition of tonal languages and regional or non-native language variants.

\newpage
\bibliographystyle{IEEEtran}
\bibliography{allophant}

\end{document}